\documentclass[12pt]{iopart}

\usepackage{graphicx}
\usepackage{bm}
\usepackage{xcolor}
\usepackage[numbers]{natbib}
\usepackage{tabularx}
\usepackage{booktabs}
\usepackage{arydshln}
\usepackage{array}
\usepackage{soul}

\newcolumntype{C}[1]{>{\centering\let\newline\\\arraybackslash\hspace{0pt}}m{#1}}

\newcommand{\real}{\mathcal{R}}

\newcommand{\synth}{\mathcal{Y}}

\newcommand{\val}{VE_{min}}
\newcommand{\graph}{GSE_{min} }
\newcommand{\mval}{$\val\,$}
\newcommand{\mgraph}{$\graph\,$}
\newcommand{\baseline}{Kustowski et al.}

\begin{document}

\title[Transformer-Powered Surrogate Design]{Transformer-Powered Surrogates Close the ICF Simulation-Experiment Gap with Extremely Limited Data}

\author{
    Matthew L. Olson\textsuperscript{1}, 
    Shusen Liu\textsuperscript{1}, 
    Jayaraman J. Thiagarajan\textsuperscript{1},  
    Bogdan Kustowski\textsuperscript{1}, 
    Weng-Keen Wong\textsuperscript{2}, 
    Rushil Anirudh\textsuperscript{1}
}

\address{
    Lawrence Livermore National Laboratory\textsuperscript{1}, 
    Oregon State University\textsuperscript{2} 
}

\ead{liu42@llnl.gov}
\vspace{10pt}

\begin{abstract}
Recent advances in machine learning, specifically transformer architecture, have led to significant advancements in commercial domains. These powerful models have demonstrated superior capability to learn complex relationships and often generalize better to new data and problems. This paper presents a novel transformer-powered approach for enhancing prediction accuracy in multi-modal output scenarios, where sparse experimental data is supplemented with simulation data. The proposed approach integrates transformer-based architecture with a novel graph-based hyper-parameter optimization technique. The resulting system not only effectively reduces simulation bias, but also achieves superior prediction accuracy compared to the prior method. We demonstrate the efficacy of our approach on inertial confinement fusion experiments, where only 10 shots of real-world data are available, as well as synthetic versions of these experiments.
\end{abstract}

\section{Introduction}
Simulation-driven science relies on the premise that sophisticated computational simulations can enable researchers to explore complex phenomena which can be challenging to explore experimentally due to prohibitive costs, time constraints, or both. Over the recent years, we have witnessed a major interest surge~\citep{hatfield2021data,nora2017ensemble,humbird2019transfer,kustowski2019transfer,kustowski2022suppressing} in leveraging such large-scale simulation data along with machine learning (ML) methodologies to drive our understanding of complex physical systems. Despite its flexibility, this approach comes with an implicit understanding that simulations are often lower fidelity representations of the true physical phenomena and can hence contain critical gaps when translating insights to real experiments~\citep{schmidt2009distilling}. In other words, ML models trained purely on simulation data can inherit its biases, and limitations, and can eventually lead to severe miscalibration with respect to the experiments.

A viable approach to mitigate this gap is to systematically adapt simulation-trained models using a handful of experimental observations, enabling the models to adjust their biases to match experimental measurements more closely through transfer learning, a method where a model developed for one task is repurposed for another~\citep{pan2009survey}. When successful, this strategy can be remarkably effective at predicting experiment outcomes (or even intermediate states) accurately, while requiring only a small fraction of the experimental observations that typically would be needed to train sophisticated ML models (e.g., deep neural networks) if experimental data alone were used~\citep{kustowski2022suppressing}. However, two critical challenges need to be addressed when building practical, transfer learning protocols: (i) the heightened risk for overfitting in cases of extremely few-shot data ($\sim$10-20 experiments), since surrogates can typically contain a large number of parameters of the order of hundreds of thousands or even millions; and (ii) the lack of clear guidance for hyper-parameter selection (e.g., learning rate, number of epochs for optimization). Imprecise choice of hyper-parameters during model fine-tuning can lead to several undesirable effects (e.g., excessive feature distortion or an increased risk of simplicity bias \cite{trivedi2023closer}), therefore resulting in poor generalization. The conventional practice of using a held-out validation dataset for hyper-parameter selection is no longer applicable to our setting of performing transfer learning with very limited data.

In this work, we address these issues using inertial confinement fusion (ICF)~\citep{betti2016inertial} as a test bed, where the simulation-experiment gap is well documented~\citep{humbird2019transfer,kustowski2019transfer} and the number of available experimental observations are very few (10) due to their high cost ($\sim $\$1M/experiment). 
First, recognizing the need for a more generalizable base model to enhance transfer learning performance, we introduce a novel framework specifically for training in transformer-based architectures~\citep{vaswani2017attention}.
Transformers have demonstrated their adaptability and effectiveness across many domains, including language \cite{devlin2018bert,radford2018improving,radford2019language,brown2020language,bubeck2023sparks}, vision \cite{dosovitskiy2021an,zhai2022scaling,khan2022transformers,fang2021you}, audio \cite{dhariwal2020jukebox,kreuk2022audiogen,borsos2023audiolm}, chemistry \cite{schwaller2019molecular,schwaller2021mapping,schwaller2021extraction,born2023regression},  and biology \cite{rives2021biological,jumper2021highly}. Building upon the versatility of transformers, we introduce a novel framework specifically designed for masked training in transformer-based architectures by utilizing masked auto-encoders~\citep{he2022masked}, where masking involves selectively hiding parts of the data to enhance model learning without labels. 
Targeted at enhancing the adaptability and expressiveness of ML models trained on simulations, this framework accommodates a variety of designer-specified masking strategies, such as forward modeling, inverse modeling, or combinations thereof. While the framework is flexible enough to support any masked modeling approach, we specifically focus on two strategies: forward modeling for predictive learning and random masking of an entire data sample. These strategies enable the model to jointly learn the complex dependencies between simulation inputs and outputs—akin to a standard surrogate model—as well as the correlations across disparate output modalities resembling modern representation learners.
Second, we introduce a novel hyper-parameter selection approach for model fine-tuning. Our approach models different hyper-parameter choices as nodes of a graph, their corresponding validation errors as the function at each node, and adopts a graph filtering strategy for reliable hyper-parameter recommendation. To demonstrate that our proposed techniques are statistically meaningful, we also show improvements using a larger, synthetic ICF dataset, where the simulation-experiment gap is artificially built by splitting the datasets along known physics parameters~\cite{kustowski2022suppressing}.

\begin{figure}[!bt]
\centering
        \includegraphics[width=0.99\linewidth]{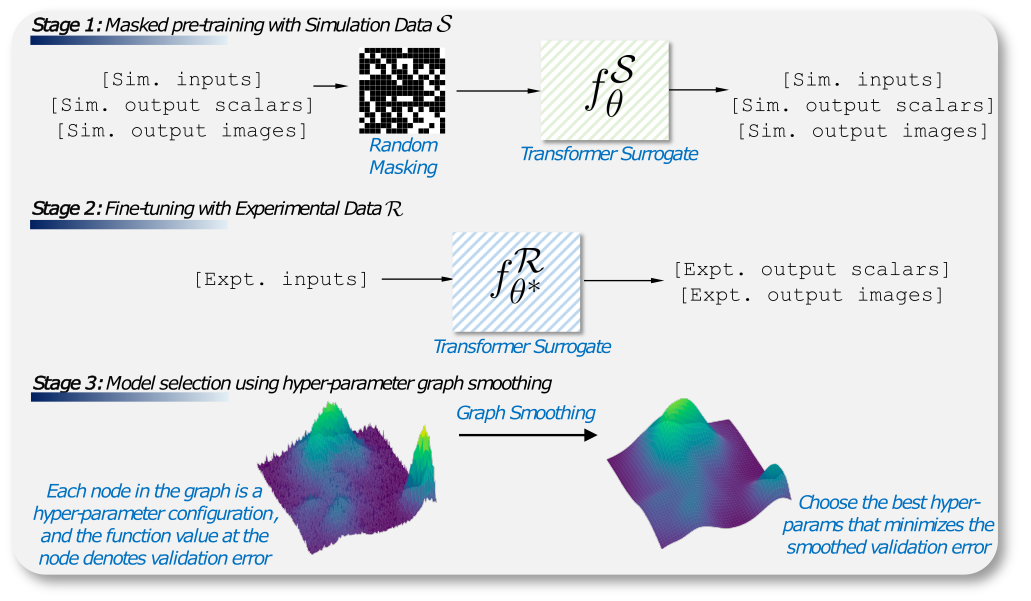}
        \caption{Our method is separated into three distinct stages: First, pretraining on simulation data with masked autoencoding and surrogate losses. Second, finetuning our model on the experimental data with a hyper-parameter sweep. Finally, finding the best hyper-parameter settings using our novel graph-based selection. }
        \label{fig:algorithm_steps}
\end{figure}

\newcommand\B{\rule[-1.2ex]{0pt}{0pt}} 

\begin{table}[tb]
    \centering
    \begin{tabularx}{\textwidth}{X l }
        Experiment Description & Reference \\
        \hline
        \parbox[t]{0.81\textwidth}{Our primary scalar prediction results showing significant improvements from our methods versus \baseline \cite{kustowski2022suppressing} \B}  & Table \ref{tbl:results} \\
        \parbox[t]{0.81\textwidth}{Our primary image predictions from our model versus \baseline \cite{kustowski2022suppressing} , with consistent improvements for the synthetic data scenario. \B}& Figure \ref{fig:image} \\
        \parbox[t]{0.81\textwidth}{A figure showing how our graph smoothing improves hyper-parameter selection. \B} & Figure \ref{fig:scatterplot_val_test} \\
        \parbox[t]{0.81\textwidth}{An analysis between pretrained learned embeddings and fine-tuned embeddings showing consistent simulation bias for simple hyper-parameter selection. \B} & Table \ref{tbl:cka} \\
        \parbox[t]{0.81\textwidth}{Experiment using significantly more synthetic data (50 training points) to show graph smoothing matches minimum validation error. \B}  & Table \ref{tbl:50synth} \\
        \hline
    \end{tabularx}
    \caption{Table of Experiments. }
    \label{tbl:experiments}
\end{table}

\paragraph{\textbf{Main Findings}}
We evaluate our methods on a real-world benchmark from the literature~\citep{kustowski2019transfer}, which comprises ICF simulations and real experiments curated at the National Ignition Facility (NIF); and a more recent Hydra simulation-based synthetic benchmark~\citep{kustowski2022suppressing} that emulates the large distribution shifts typically observed in the real world. We find that our transformer-based surrogate, combined with our robust hyper-parameter selection strategy, is significantly more effective at bridging the simulation-experiment gap, offering a relative gain of $\sim 40\%$ in terms of predictive error over the state-of-the-art neural network surrogates. More specifically, we find that our richer class of transformer-based surrogates enables us to employ a simpler transfer learning protocol (a simple linear-bias correction as opposed to extensive neural network weight fine-tuning), therefore, making it ideal for applications operating in very small experimental data regimes. Next, we find that the graph-based hyper-parameter selection strategy yields much more robust and generalizable models that outperform traditional validation techniques significantly. 
We present an overview of our method in figure \ref{fig:algorithm_steps}, and we summarize our experiments in table \ref{tbl:experiments}.

\section{Experimental Setup and Results}


In our effort to bridge the gap between simulation and experimental data, we employ our proposed framework integrating masked training in transformer-based architectures and a graph-based hyper-parameter selection strategy that is particularly effective when the number of experimental observations is very small. We begin by assessing the framework's performance on the inertial confinement fusion (ICF)~\cite{atzeni2004physics,betti2016inertial} datasets, which present substantial challenges due to limited availability and high costs of experiments. To demonstrate the effectiveness of our proposed approaches, we build upon the work of \citet{kustowski2022suppressing} by using the benchmarks presented in their study. 
\vspace{5pt}

\noindent \textbf{Datasets:} Specifically, we use two datasets in our experiments: The first, referred to as $\real$, stems from real inertial confinement fusion (ICF) experiments conducted during a ``Bigfoot'' campaign in 2018~\cite{casey2018high} at the National Ignition Facility (NIF) in Livermore, California. This multi-modal dataset comprises 10 ICF shots and is accompanied by a large set of simulations produced using a 1D physics simulator \cite{kustowski2022suppressing}, denoted as $\mathcal{S}$. The dataset consists of nine scalar inputs corresponding to the design space of the simulator and experiments, ten output scalar values, and an output X-ray image.
Most of the inputs relate to the laser energy's conversion into X-rays and its impact on capsule compression, including energy, power, and geometric asymmetry; the other inputs concern hydrodynamic scaling, fuel preheat, and capsule material properties.
The 10 scalar outputs capture key phenomena such as the precise moments of peak neutron and X-ray emissions, referred to as ``bang times'', alongside essential thermodynamic variables like temperature and velocity. Additionally, the dataset includes detailed profiles of X-ray emissions and neutron yields, the latter being a critical indicator of the experimental yield. The overarching aim is to enhance our predictive capabilities, thereby enabling us to maximize the experimental energy yield. 
The second dataset, denoted as $\synth$, was generated from a multi-modal surrogate \cite{anirudh2020improved}, previously trained on all the aforementioned simulations. The domain shift here is synthetically induced by obtaining predictions from the surrogate across a disjoint set of input parameters \cite{kustowski2019transfer}. This allows us to test our hypothesis on a much larger set of data (1000 samples in total) to obtain more statistically significant results. Even with the synthetic set, we always assume access only to a very few number of samples for fine-tuning, but here we can use a much larger test set for evaluations, following \citet{kustowski2019transfer}'s protocol.

\vspace{5pt}

\begin{table}[tb]
\centering
\begin{tabular}{lC{0.95in}C{0.5in}C{0.95in}C{0.5in}}
\texttt{Scalar ID. Name}                    & \footnotesize{Kustowski~et~al.} $(\real )$ & Ours $ (\real)$ & \footnotesize{Kustowski~et~al.} $(\synth )$  & Ours ($\synth$)  \\ \toprule
&  \multicolumn{2}{c}{\emph{Leave-One-Out Setting}}  & \\

 &  &  &  & \\
\texttt{1. Neutron bang time}              & 0.243           & \textbf{0.037}     & 0.804           & \textbf{0.664}  \\
\texttt{2. X-ray bang time}                & 0.267           & \textbf{0.029}     & 1.037           & \textbf{0.679}  \\
\texttt{3. Downscattered ratio}            & 0.920           & \textbf{0.550}     & 5.490           & \textbf{4.495}  \\
\texttt{4. Temperature}                    & 0.233           & \textbf{0.152}     & 4.351           & \textbf{2.893}  \\
\texttt{5. Hot spot radius}                & 0.130           & \textbf{0.116}     & 9.059           & \textbf{6.788}  \\
\texttt{6. Velocity}                       & 0.321           & \textbf{0.212}     & 8.615           & \textbf{6.970}  \\
\texttt{7. X-ray emission}                   & 1.363           & \textbf{0.745}     & 8.262           & \textbf{4.516}  \\
\texttt{8. Neutron yield}                & 0.058           & \textbf{0.035}     & 8.389           & \textbf{4.355}  \\
\texttt{9. Neutron burn width}              & 0.404           & \textbf{0.320}     & 9.030           & \textbf{8.851}  \\
\texttt{10.\hspace{-0.5em} X-ray burn width}             & 4.758           & \textbf{2.728}     & \textbf{10.770}           & 11.342 \\
 &  &  &  & \\

Scalars (avg. of above)         & 0.870           & \textbf{0.492}     & 6.580           & \textbf{5.160}    \\ 
Images                          & 0.170           & \textbf{0.154}     & 0.079             & \textbf{0.030}  \\ 
\bottomrule
 &  \multicolumn{2}{c}{\emph{Leave-3-Out Setting}}  & \\
 &  &  &  & \\
Scalars (avg.) & 73.438 & \textbf{0.631} & 7.974 & \textbf{7.255} \\
Images         & 1.445  & \textbf{0.189} & 0.089 & \textbf{0.055} \\ 
\bottomrule

\end{tabular}
\caption{The average MSE over all leave-one-out test samples using our graph optimized model, compared to the baseline, on both the simulated and experiment datasets. Our model often has large performance increases over the baseline for both scalar predictions and image predictions.
}
        \label{tbl:results}
\end{table}


\noindent \textbf{Evaluation metrics:} To assess the efficacy of our proposed methods, we use the Mean Square Error (MSE) as the primary evaluation metric for both scalar and image-based predictions. Each experimental setup was executed 10 times, with a leave-one-out cross-validation across the 10 available data samples in the real dataset. In each cross-validation fold, one sample is used for testing, one for validation, and the remaining 8 for fine-tuning. For consistency, we use the same setup in the synthetic dataset (8 train, 1 validation) during fine-tuning and model selection but increase the test set to all the remaining available samples (991). This is repeated 10 times, with the train and val data chosen at random without replacement.

\noindent \textbf{Results:} The aggregated results are presented in Table \ref{tbl:results} (top). Our approach is compared against baseline methods on both experimental and synthetic datasets. Across the board, our method demonstrates a substantial reduction in the MSE values for both scalar and image predictions. Specifically, on the experimental dataset, we observed an average reduction of nearly 50\% in the MSE, declining from 0.87 to 0.492. For the synthetic dataset, the error rate decreased from 6.580 to 5.155, nearly a $20\%$ improvement.

For a more comprehensive evaluation, we also conducted additional experiments with seven training data points, as shown in Table \ref{tbl:results} (bottom), aligning with the experimental setup described in \citet{kustowski2022suppressing}. In this setting, we trained models using all possible combinations of seven data points, leading to a total of 120 individual experiments. The performance degraded slightly when utilizing fewer training samples, as expected, but our proposed method still significantly outperformed the baseline, exhibiting remarkable gains in predictive accuracy for both scalar and image outputs.

\paragraph{Comparative Statistical Evaluation of Hyper-parameter Selection Strategies}
For additional experimental evaluation, we use our ``leave-3-out'' experiments for further statistical analysis shown in Table \ref{tbl:results} (top). It is evident that our proposed method consistently outperforms the baseline algorithm. However, to offer a quantitative comparison, we focus on contrasting our Minimum Smoothed Error Graph (hereinafter denoted as $\graph$) with the Traditional Minimum Validation Error ($\val$). A detailed table for the leave-3-out experiment results (and results for leave-one-out with $\val$) can be found in the supplement. 

To ascertain the statistical significance of the performance differences between \mgraph and $\val$, we conducted a series of paired-sample t-tests. For the Mean Squared Errors (MSE) averaged over scalars, the test yields $\mu_1 = 1.027$, $\mu_2 = 0.631$, $t = 2.3134$, and $p = 0.0108$, confirming the superiority of \mgraph at a 95\% confidence level. Similarly, for the average pixel-wise MSE, we find $\mu_1 = 0.208$, $\mu_2 = 0.189$, $t = 2.0124$, and $p = 0.0227$, which again corroborates the enhanced performance of \mgraph. While the small sample size is relatively small, we emphasize the thoroughness of our approach in partitioning the dataset into all possible configurations, thereby enhancing the reliability of our statistical inferences.

\begin{figure}[!bt]
\centering
        \includegraphics[width=0.99\linewidth]{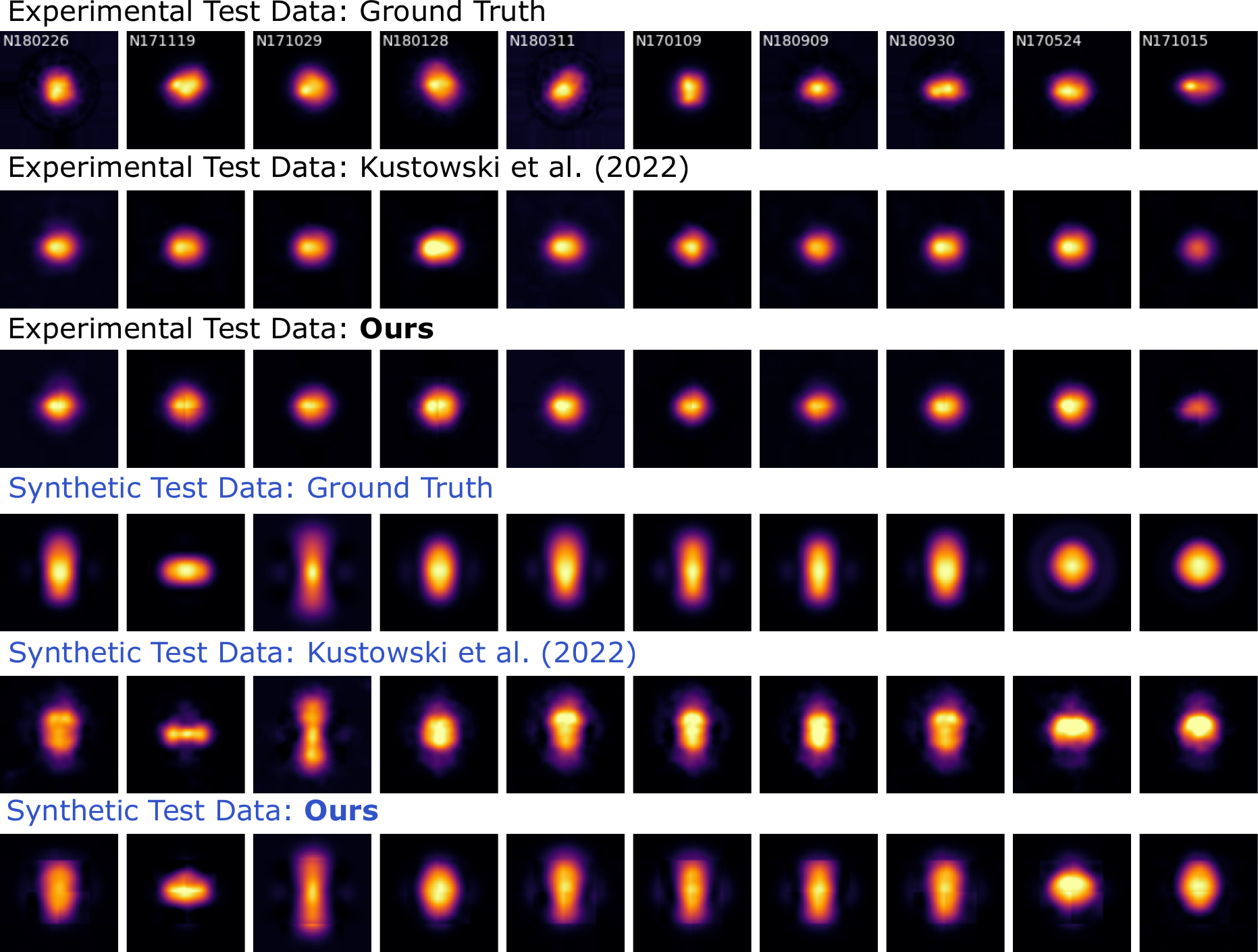}
        \caption{Our models' predictions on the held-out test X-ray images after fine-tuning on the real training data, compared to the baseline. Pixels here represent energy outputs of the experimental implosion. White pixels are high energy, purple are lower energy, and black are no energy.  Zoom in to better see the results. The MSE for our method is lower than the baseline for the test predictions. While the image quality is not perfect, we find that our new model has modest improvements over the baseline both qualitatively and quantitatively.}
        \label{fig:image}
\end{figure}

\paragraph{Diagnostic X-ray Images}
We commence our discussion with an analysis of the model's efficacy on the reconstructed images, as depicted in Figure \ref{fig:image}, in comparison to the baseline method. Our model exhibits a superior ability to approximate the underlying distribution of the training set. In particular, we draw attention to the synthetic image results, which demonstrate a marked reduction in simulation bias in our approach.

Although our generated images display minor artifacts attributable to the use of transformer-based patching techniques, they successfully approximate the overarching geometric structures. It is crucial to note the primary focus of our study lies not in image reconstruction but in the accurate prediction of scalar values. Our dataset is multimodal, comprising diagnostic images and scalar values; however, the latter serve as the principal targets of interest. The notable improvement in the prediction of these scalar attributes for the experimental dataset underlines the practical significance of our approach.

\begin{figure}[!bt]
\centering
        \includegraphics[width=0.99\linewidth]{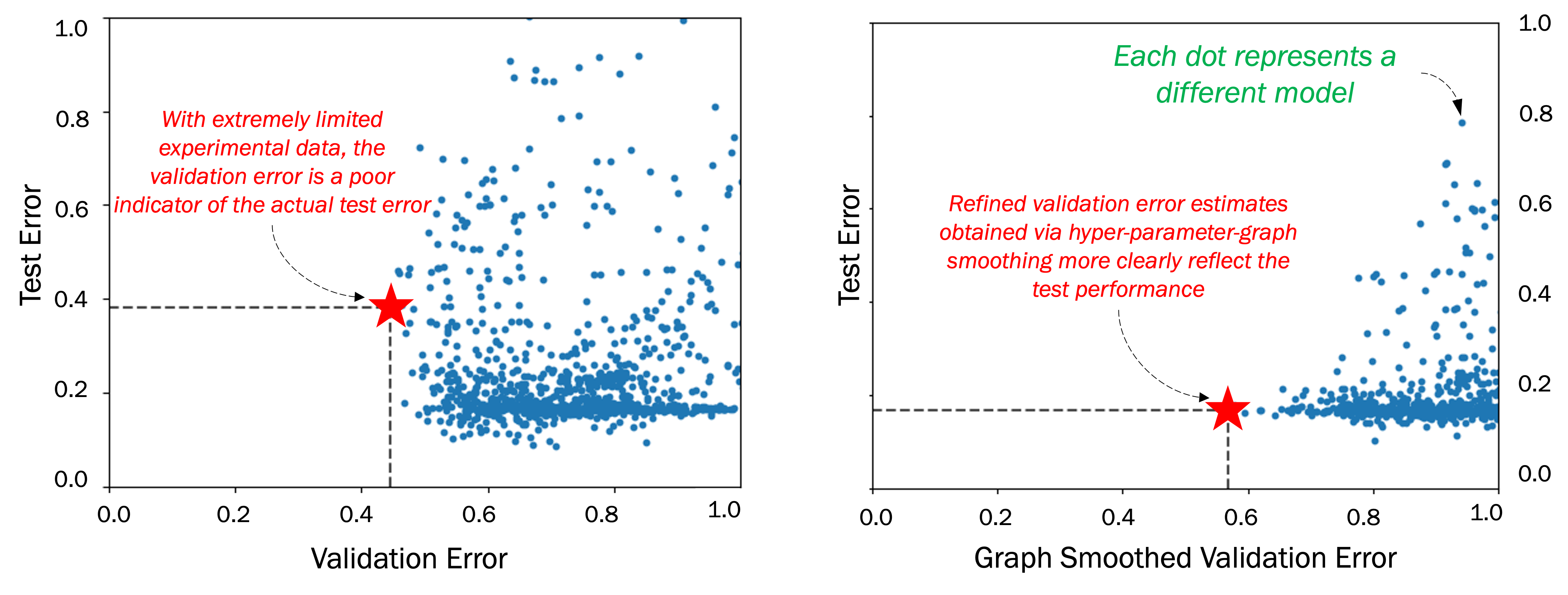}
        \caption{Hyper-parameter graph smoothing ensures optimal model selection based on noisy validation error. \textbf{(Left:)} Validation versus test error rates for scalar predictions experiment data. \textbf{(Right):} Proposed graph-smoothed validation error vs test error. We highlight the minimum validation error and the minimum smoothed validation error finding that our smoothing removes the noisy data to find a robust, well performing hyper-parameter selection.}
        \label{fig:scatterplot_val_test}
\end{figure}
\paragraph{Robust Hyper-parameter Optimization via Graph Smoothing}
Figure \ref{fig:scatterplot_val_test} elucidates the efficacy of our graph smoothing hyper-parameter optimization method, elaborated in Section \ref{sec:hpo}. The primary utility of this method lies in its ability to remap instances characterized by a disparity between validation and test errors into a refined validation error space. By applying this smoothing operation, we uncover regions within the hyper-parameter landscape that robustly yield low test errors.

The figure plots validation against test errors for multiple hyper-parameter configurations, thereby empirically demonstrating the algorithm's robustness. Notably, configurations that initially exhibit high test errors, despite low validation errors, are effectively smoothed out. This results in a more reliable selection of well-performing hyper-parameters, as evidenced by the sparsity of such points in the modified validation space.

\begin{figure}[!bt]
\centering
        \includegraphics[width=0.99\linewidth]{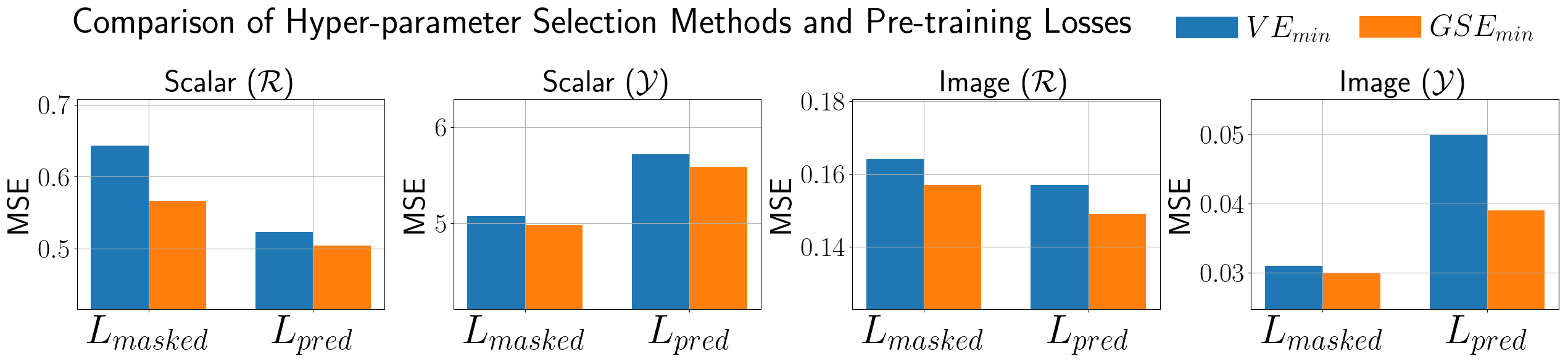}
        \caption{Detailed results comparing the masking strategies $L_{masked}$ and $L_{pred}$, as well as using the smoothed graph validation error rate \mgraph versus the non-graph minimum validation error rate \mval. We find the interesting result: masking is useful for the $\synth$ dataset, but not for $\real$. Furthermore, using the graph is always an improvement. 
        }
        \label{fig:mask_vs_pred}
\end{figure}
While our primary results are consistent improvements over the baseline, we take a deeper look at how our pretraining losses affect the results of our models. We compare two pretraining strategies. The first is forward surrogate modeling prediction loss ($L_{pred}$): only predict simulation outputs given simulation inputs. The second is forward loss in addition to masked auto-encoding loss ($L_{masked}$): the model randomly sees partial inputs and partial outputs and, then, predicts what it does not see. Furthermore, we take a detailed look at how \mgraph and \mval perform when separately analyzing the losses. 

The graphs in figure \ref{fig:mask_vs_pred} provide insights into the relationship between hyper-parameters and model performance, showing an interesting behavior for the use of the masked auto-encoding loss. We see that on the synthetic dataset $\synth$, the use of masking is a large improvement over the pure prediction loss. We also find the opposite to be true for $\real$. This is most likely due to the former dataset’s size: It is a much smaller distribution shift between the pretraining dataset and the fine-tuning dataset, such that the learned correlations from $L_{masked}$ can easily be accounted for, whereas the changes in $\real$ are so dramatic that deeper correlations learned from $L_{masked}$ result in overfitting.
We also highlight that for all our experiments shown in Figure \ref{fig:mask_vs_pred}, using the smoothed graph validation errors, \mgraph consistently results in enhanced performances for all experiments over simply using the minimum validation error \mval.

\begin{table}[htbp]
\centering
\begin{tabular}{ p{1.7in} c c c }
Scalar Name            & \footnotesize{Kustowski~et~al.} $(\synth )$ & $\val$ & $\graph$ \\ \toprule
\texttt{Neutron bang time}      & 0.595{\scriptsize$\pm 0.153$ }& \textbf{0.219}{\scriptsize$\pm 0.036$ }& 0.223{\scriptsize$\pm 0.038$ }\\ 
\texttt{X-ray bang time}        & 0.926{\scriptsize$\pm 0.141$ }& \textbf{0.218}{\scriptsize$\pm 0.044$ }& 0.222{\scriptsize$\pm 0.046$ }\\ 
\texttt{Downscattered ratio}    & 5.085{\scriptsize$\pm 0.673$ }& 0.638{\scriptsize$\pm 0.247$ }& \textbf{0.634}{\scriptsize$\pm 0.195$ }\\ 
\texttt{Temperature}            & 3.139{\scriptsize$\pm 0.244$ }& 0.454{\scriptsize$\pm 0.147$ }& \textbf{0.442}{\scriptsize$\pm 0.153$ }\\ 
\texttt{Hot spot radius}        & 7.121{\scriptsize$\pm 0.905$ }& 0.929{\scriptsize$\pm 0.233$ }& \textbf{0.908}{\scriptsize$\pm 0.218$ }\\ 
\texttt{Velocity}               & 6.047{\scriptsize$\pm 0.657$ }& \textbf{0.479}{\scriptsize$\pm 0.191$ }& \textbf{0.479}{\scriptsize$\pm 0.187$ }\\ 
\texttt{X-ray emission}         & 6.663{\scriptsize$\pm 0.471$ }& 0.547{\scriptsize$\pm 0.261$ }& \textbf{0.542}{\scriptsize$\pm 0.249$ }\\ 
\texttt{Neutron yield}          & 7.029{\scriptsize$\pm 0.673$ }& \textbf{0.504}{\scriptsize$\pm 0.211$ }& 0.511{\scriptsize$\pm 0.201$ }\\ 
\texttt{Neutron burn width}     & 8.141{\scriptsize$\pm 0.705$ }& 2.080{\scriptsize$\pm 0.549$ }& \textbf{2.053}{\scriptsize$\pm 0.463$ }\\ 
\texttt{X-ray burn width}       & 8.595{\scriptsize$\pm 0.587$ }& \textbf{2.900}{\scriptsize$\pm 0.581$ }& 2.933{\scriptsize$\pm 0.630$ }\\ 
 &  &  &  \\ 
Scalars (avg.) & 5.334{\scriptsize$\pm 0.189$ }& 0.897{\scriptsize$\pm 0.840$ }& \textbf{0.895}{\scriptsize$\pm 0.844$ }\\ \midrule
Images                 & 0.066{\scriptsize$\pm 0.005$ }& \textbf{0.005}{\scriptsize$\pm 0.002$ }& \textbf{0.005}{\scriptsize$\pm 0.002$ }\\ \bottomrule
\end{tabular}
\caption{\textbf{Graph smoothing converges to standard model selection when more data is available}: Here, we use 50 synth train and 10 validation examples. Once again, our method outperforms the baseline significantly. Reassuringly, we note that with increased availability of training and validation data, our \mgraph approach converges to standard model selection based on minimal val error \mval.}
\label{tbl:50synth}
\end{table}
\paragraph{Effects of Increased Training Data}
In an effort to understand the model's performance in data-rich scenarios, we conduct an ablation study utilizing 50 data points for fine-tuning, as presented in Table \ref{tbl:50synth}. As the definition of "few-shot" learning can be ambiguous in the literature, we consider the scenario with 50 points to not be few. Nevertheless, our findings indicate that both \mval and \mgraph yield comparable performance, significantly surpassing the baseline. This suggests two critical insights: First, our transformer-based model consistently outperforms the non-transformer baseline. Second, in scenarios with cleaner, less noisy validation data, the graph smoothing operation poses no detriment to model performance. 

Another effect of additional data is a change in the optimal hyper-parameters. We compare the hyper-parameter configurations selected by all runs between the data-scarce experiments and this relatively data-rich experiments. Our analysis revealed a degree of consistency in hyper-parameters across different data scales, such as identical learning rates and a high number of training epochs. However, variations were observed in the selection of fine-tuning layers and other hyper-parameters, showing the importance of validation metrics within a dataset.

\paragraph{Extreme Case: One-Shot Learning}
To explore the limitations of our method, we conducted an experiment with only one data point for training and another for validation. As anticipated, the results are markedly sub-optimal; however, the performance of \mval and \mgraph is indistinguishable in this extreme setting. This results serves to corroborate that \mgraph essentially reduces to \mval when the data becomes extremely sparse.



\begin{table}[!bt]
    \centering
    \begin{tabular}{ l p{0.35in} p{0.35in} p{0.35in} p{0.35in} p{0.35in} p{0.35in} p{0.35in} p{0.35in} p{0.35in} p{0.35in} }
    Scalar ID & 1 & 2 & 3 & 4 & 5 & 6 & 7 & 8 & 9 & 10 \\
    \hline
    \mgraph &  \textbf{0.518} & \textbf{0.493} & \textbf{0.431 }& \textbf{0.462} & \textbf{0.561 }& \textbf{0.501} & \textbf{0.460} & \textbf{0.527 }& \textbf{0.517 }& \textbf{0.521} \\
    \mval & 0.527 & 0.507 & 0.444 & 0.470 & 0.579 & 0.504 & 0.463 & 0.542 & 0.525 & 0.524 \\
    \hline
    \end{tabular}
    \caption{Using CKA to compare the similarity between pretrained feature embeddings and finetuned feature embeddings from one left-out test point. We show how our proposed method for \mgraph results in less simulation bias (lower similarity score compared to the pretrained  embedded features) for all scalar embeddings when compared against using \mval. }
    \label{tbl:cka}
\end{table}

\subsection{Analysis of feature Embeddings using CKA}
Centered Kernel Alignment (CKA) is a technique used to measure the similarity between two sets of features \citep{kornblith2019similarity}. It has been widely utilized in the context of neural network representations to understand the alignment of features in different layers or networks. In short, it gives you a similarity between two distributions of features. If the features are identical, then the score will be $1.0$; the more the features' distributions deviate, the lower the score will go towards zero. Here we use CKA to compare the features between our two hyper-parameter selection strategies \mval and $\graph$.

In Table \ref{tbl:cka}, we show the results of our CKA analysis across the embeddings for all 10 output scalars from our leave-one-out experiments. Specifically, we apply CKA to compare the embeddings from pretrained embeddings to embeddings from $\val$, and pretrained embeddings to $\graph$.  Our analysis demonstrates a clear pattern: the use of \mgraph embeddings is consistently lower than the \mval embeddings. These lower scores indicate that \mgraph consistently exhibits less simulation bias as compared to the embeddings obtained from \mval. 

It is important to note the limitations of CKA, as discussed in recent literature \citep{davari2023reliability}, that performance can be influenced by outliers. This sensitivity to outliers implies that while CKA scores provide a useful comparative measure of feature similarity, they should be interpreted with caution. The observed differences in CKA scores, particularly those of minimal magnitude, should be considered indicative of a broader trend towards reduced similarity with the pretrained model rather than definitive evidence of the superiority of one method over another. Our findings suggests that the graph-based method might be a more robust and unbiased approach for generating embeddings in our context.



%

\section{Discussion}
In the current study, we advance the field of few-shot transfer learning in scientific contexts by introducing a novel approach that harnesses the versatility of Transformer-based architectures. Extending this versatility, our model is uniquely equipped to handle multi-modal data, incorporating both scalar and image formats seamlessly. This capability enables the model to predict complex physical systems with significantly less simulation bias.

A crucial part of our strategy is the innovative graph-based hyper-parameter optimization technique. Previous studies have explored few-shot learning and hyper-parameter optimization from different angles. For instance, \citet{franceschi2018bilevel} introduced a bilevel programming framework for gradient-based hyper-parameter optimization and meta-learning, particularly for deep learning and few-shot learning scenarios. On the other hand, \citet{mazumder2021rnnp} developed a robust few-shot learning approach without specifically focusing on hyper-parameter optimization.

In contrast, while \citet{van2018hyperparameter} analyzed the importance of various hyper-parameters, they did not factor in the challenge of untrustworthy validation data, which our work addresses. \citet{Liang_2022_CVPR} also recognized the issue of noisy labels in few-shot learning but diverged by choosing to incorporate sophisticated loss functions rather than emphasizing hyper-parameters. Our method, countering traditional challenges such as noisy validation error rates seen in prior work, leads to more reliable and generalizable hyper-parameter configurations that improve overall model performance. Furthermore, \citet{muniraju2020coverage} presented parameterized coverage-based designs for superior sample mining and hyper-parameter optimization, indicating the increasing significance of these concepts in the scientific community.

Beyond optimization, our study's emphasis on surrogate modeling and addressing simulation bias stands on the shoulders of substantial previous research. Surrogate modeling, for example, has seen applications in varied scientific domains, from the rigorous optimization framework for expensive functions used in helicopter rotor blade design by \citet{booker1999rigorous} to Bayesian calibration techniques for computer models introduced by \citet{kennedy2001bayesian}. In the specific arena of Inertial Confinement Fusion (ICF), the field has witnessed machine learning-driven efforts like that of \citet{hatfield2021data}, ensemble models from \citet{nora2017ensemble}, and neural network-based approaches such as those by \citet{kustowski2019transfer} and \citet{kustowski2022suppressing}. These underline the persistent pursuit to address simulation bias and provide robust models, aligning with our work's objectives.

Building upon these foundations, our work further explores the frontier of predictive modeling within the ICF domain. A critical aspect of this exploration is the acknowledgment of potential radical changes in physical behavior in parts of the design space that remain unexplored experimentally. One such phenomenon, ignition, occurs when the energy generated within the fusion fuel surpasses the energy being lost, leading to a self-sustaining fusion reaction. This represents a drastic shift in the system's response and poses significant challenges for predictive modeling. The complexity of predicting events like ignition, particularly with simulation-based data, highlights the nonlinear and high-stakes nature of these transitions. Our approach, designed to enhance the predictive model's capability across a broad spectrum of conditions, aims to contribute to a more comprehensive understanding and optimization of experimental yields in ICF research. By addressing these challenges, we pave the way for breakthroughs in fusion energy.

What sets our work apart is its potential for facilitating multi-modal transfer learning tasks in scientific domains. While the immediate impact of our contributions is evident, this work also lays the groundwork for more expansive research. Future sections will delve into the possibility of applying our methods to other disciplines, thereby widening the scope and impact of our findings.

\section{Methods}
\subsection{Formal Definitions}
We consider multi-modal physics simulation datasets given by $\mathcal{D}^s = (\mathcal{X},\mathcal{O},\mathcal{I})$  consisting of input scalars $\mathcal{X} = \{x_1, x_2, \dots, x_N\}$, output scalars $\mathcal{O} = \{o_1, o_2, \dots, o_N\}$, and output images $\mathcal{I} = \{I_1, I_2, \dots, I_N\}$, where $N$ denotes the size of the dataset and $\mathbf{d_j} = (x_j,o_j,I_j)$. We also assume access to a ``target'' dataset $\mathcal{D}^t$ which is ultimately the domain on which we want our model to be most accurate. We expect $\mathcal{D}^s\neq\mathcal{D}^t$, due to the known gap between them. Here, the source domain is typically a simulation dataset collected by sampling from a physics simulator, and the target dataset contains real experimental observations. Consequently, we assume that the number of available target samples is very small, $N^s>>N^t$. We use the superscript notation to denote the domain (source vs target) as required, and drop it otherwise for simplicity of notation.




\paragraph{Problem Setup} Let us define a surrogate as $f_{\theta}^s:\mathcal{X}^s\rightarrow (\mathcal{O}^s, \mathcal{I}^s)$, where $\theta$ are its parameters to be learned. Due to the expected simulation-experiment gap, this model will likely perform poorly when tested directly on target data, i.e.,  we expect a large error in the prediction since $f_{\theta}^s(x^t) \neq (o^t,I^t)$. This gap typically manifests as a task shift, i.e., where the input distribution $\mathcal{X}$ remains unchanged but the output distribution has changed significantly between source and target. As a result, the source model must be adapted or fine-tuned using a small number of training examples from $\mathcal{D}^t$ so that this gap can be closed. 

\paragraph{Fine-tuning and model adaptation}

The biggest challenge in model adaptation in this context is the lack of sufficient training data. This makes the fine-tuning problem challenging due to two main reasons:

(i) \textbf{Risk of overfitting} -- While increasingly complex models with a large number of parameters can provide more useful inductive biases to ML surrogates, fine-tuning all the parameters on a very limited dataset will likely result in overfitting. To mitigate this issue, only a part of the network is adapted (typically the final few layers, though not always) while keeping the rest of the parameters fixed. In other words, we can split the parameters as $\theta^s = [\beta^s_{\mathrm{fixed}},\beta^s_{\mathrm{trainable}}]$, indicating weights that are unchanged and weights that get updated. The fine-tuned model is typically of the form $\theta^* = [\beta^s_{\mathrm{fixed}},\beta^*_{\mathrm{trainable}}]$, where $^*$ indicates the final, fine-tuned model that is used to make predictions. 

(ii) \textbf{Model selection with less validation data} -- Model selection is the problem of identifying the best set of hyper-parameters based on the performance on a held-out validation set (not seen during training). When the validation set is very small -- as is likely the case when available labeled data for fine-tuning itself is very sparse -- the best performing model on the validation set is unlikely to be the best performing model on the real test, due to very noisy estimates arising from very poor sampling of the validation set. As such, picking a model that is likely to generalize well is very challenging.

In the methods section, we outline our solution to both of these problems and show how the proposed transformer-based surrogate and model selection strategy are effective in addressing the simulation-experiment gap.



\subsection{Masked training with Transformer Surrogates}
Our first, and one of two main contributions, is the use of transformer models \cite{vaswani2017attention} as surrogates in the ICF application space. Transformers are a class of general-purpose learners that work on tokenized forms of data (such as patches or chunks) and learn attention across arbitrary data modalities \cite{radford2021learning}. This enables them to capture important correlations on their own and, equally important, the architecture makes very few assumptions about the data. This phenomenon has led to successes in a variety of applications, such as computer vision \cite{dosovitskiy2021an} and other multi-modal data \cite{li2019neural}. In particular, we explore the use of masked training in transformers using the Masked Auto-Encoder (MAE) \cite{he2022masked}. Inspired by the successes of masked pre-training in language modeling, the MAE presents a pre-training strategy for image data that was a significant breakthrough in self-supervised representation learning for image data. We extend the MAE strategy from just one modality (text or image) to multiple modalities.

\begin{figure}[!bt]
\centering
        \includegraphics[width=0.99\linewidth]{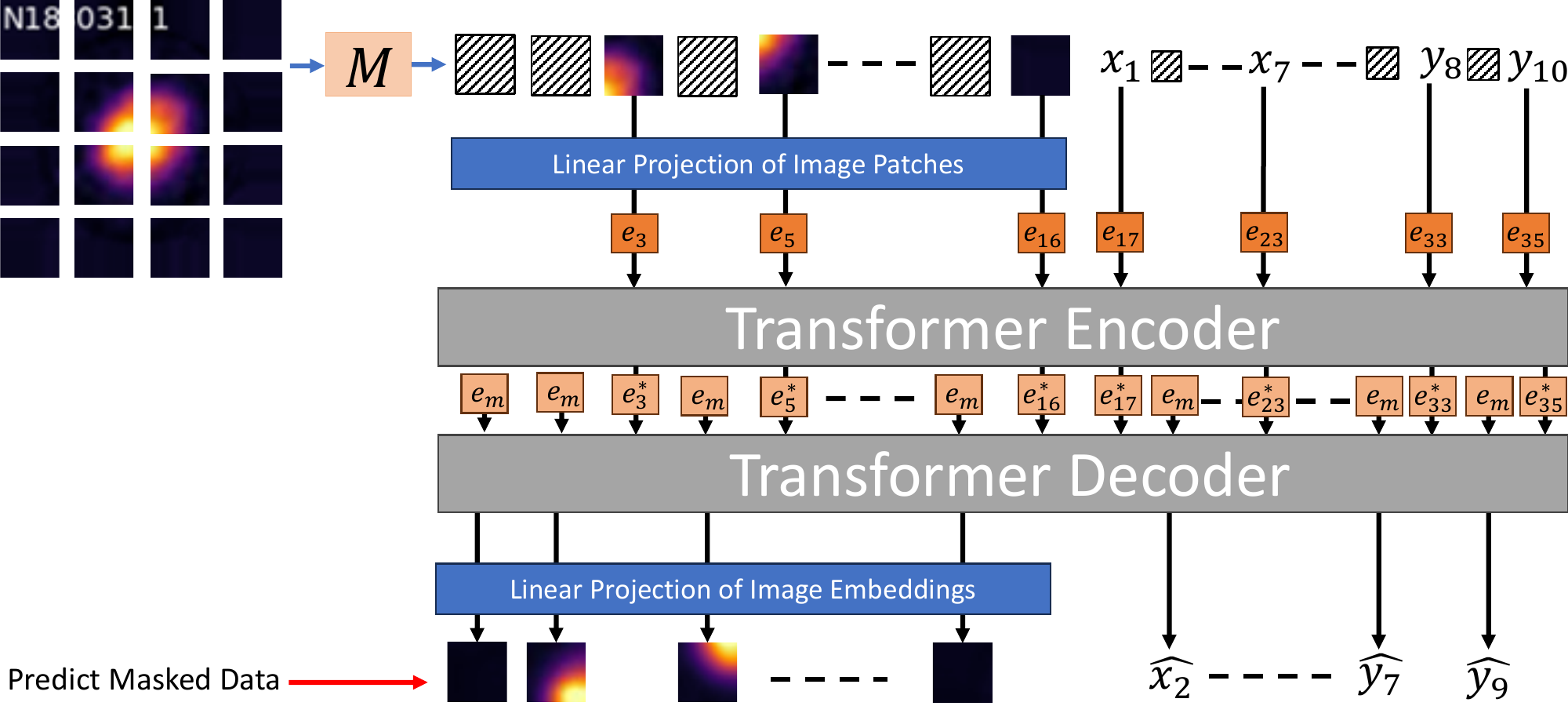}
        \caption{\textbf{Masked Pre-training:} Our novel multi-modal architecture leverages both images and scalars as inputs and outputs for a transformer-based deep neural network. Transformers enable straightforward surrogate models as well as effective representation learning through masked autoencoding. 
        }
        \label{fig:architecture}
\end{figure}

In order to effectively leverage masked autoencoding, we have to employ a deep transformer-based model.
A diagram of our model is shown in figure \ref{fig:architecture}.

\paragraph{Generalized Surrogate Model with Flexible Masking Strategies}

While a traditional surrogate model is often defined as $(o_j,I_j) = f_{surr}(x_j)$, in this work we explore a new formulation in order to capture correlations. Prior methods are designed around learning a representation 
 that captures the correlations 
 between $\mathcal{Y}$ and $\mathcal{I}$ by learning a compressed representation jointly. However, in this work, by utilizing a deep transformer-based neural network, we can effectively capture these correlations, in addition to including $\mathcal{X}$ in the learned representation. Therefore, we introduce a more general version of $f$ by incorporating multiple strategies from our novel masking framework which we define as follows. Let $\mathcal{M} = (M_{forward}, M_{random})$ be a set of masking functions, each of which takes as input a data sample $d_j$ and returns only some element, i.e., $o_j, I_j = M_{forward}(d_j)$ corresponding to a standard forward surrogate model $o_j, I_j = M_{forward}(d_j) = f_{surr}(x_j)$.
 We also note the inverse of a mask $\bar{M}$ to be the opposite of said mask, i.e., $x_j = \bar{M}_{forward}(d_j)$. Our other masking strategies $M_{random}$ randomly selects from all elements of a data sample to mask at a fixed random rate ($75\%$ in our case). We emphasize that while our task only requires the two masking strategies, other strategies can be defined for other data representation tasks (such as an inverse mask), hence the flexibility of our framework. 

The general model $f_{\theta}^s$ is a deep transformer-based neural network that can take as input all scalars and images, correspondingly masked by a desired mask $M$, then, outputs all scalars and images for a given sample $j$:

\begin{equation}
(\hat{x_j},\hat{o_j},\hat{i_j}) = f(M(d_j))
\end{equation}

The mask enables flexible training of either: a standard surrogate style with only output prediction by using mask $M_{forward}$, or for standard masked auto-encoding training where inputs are randomly selected to be masked $M_{random}$.

\begin{enumerate}
     
\item We convert our data into embeddings, as all transformer-based operations deal with embeddings rather than raw data.
\item We encode the scalars into an embedding by simply multiplying a trainable embedding by the normalized (0-1) scalar.
\item We follow standard practice \cite{dosovitskiy2021an} by flattening image patches and learning a shared image embedding space by multiplying each patch by a learnable matrix $W_p$. 
\item For each embedding we add a positional encoding embedding. The image embeddings get a fixed 2d-sinusoidal encoding, whereas the scalars get a simple trainable encoding added.
\item Our transformer model is split into two parts: the encoder and the decoder.
\item Each part is comprised of multiple transformer layers: Multi-Head Self-Attention, Layer Normalization \cite{ba2016layer}, and a Feed-forward Neural Network.
\item The outputs of the encoder are combined with a series of mask tokens embeddings, depending on the masking strategy, and are fed into the decoder network. 
\item The output of the decoder are prediction embeddings corresponding to all the data. These embeddings are either multiplied by an individual learnable prediction vector (for scalars) or by a shared prediction matrix (for images).
\end{enumerate}

During both masked and surrogate forward passes, only the available data are embedded for the encoder to process. After being encoded, a ``missing'' data embedding is placed in the location of all the missing data. This embedding has a new positional encoding added to it (still fixed for the image embeddings). All those embeddings are passed through the decoder transformer layers to get output embeddings. A learnable inverse transformation is done on all the image patches, and each scalar has its own output embedding $e_k$ and a learnable output embedding vector map (e.g., $\hat{y}_j = W_k * e_k $).

\subsection{Simulation Pretraining}
We investigate training our surrogate through two types of pretraining losses based on output prediction and masked prediction. The output prediction loss is a standard $L2$ loss on the outputs of a given example $j$ when using $M_{forward}$:
\begin{equation}
\label{eqn:Lpred}
L_{pred} = \gamma_o || \hat{o_j} - o_j ||^2_2 + \gamma_i || \hat{i_j} - i_j ||^2_2
\end{equation}

where $\gamma_i$ is a hyper-parameter tuned on the validation set of $\mathcal{S}$ and $\gamma_o = 1$.

For masking loss, we convert the image into 16 equally-sized square embeddings, along with 19 scalar embeddings. We then remove $75\%$ of those embeddings from the input to $f_{\theta}^s$ using $M_{random}$ and predict the values of the masked inputs, resulting in a mask loss defined as:
\begin{equation}
L_{masked} = || \bar{M}_{random}(x_j,y_j,i_j) - f_{\theta}^s(M_{random}(d_i)) ||^2_2
\end{equation}

The overall pretraining loss combines the prediction loss and masked auto-encoding loss, controlled by a hyper-parameter $\alpha$:

\begin{equation}
L = \alpha L_{pred} + (1 - \alpha) L_{masked}
\end{equation}

Here, $\alpha$ is a hyper-parameter tuned only on the simulation dataset. We found that setting $\alpha = 0$ (corresponding to no prediction loss) produces consistently poor results during the fine-tuning stage. And we also found $\alpha = 1$ to have inconsistent results, and therefore we treat $\alpha$ as a hyper-parameter passed down to our fine-tuning (either $\alpha = 1$ or an optimized $\alpha$ of $0.02$).

\subsection{Experimental Data Fine-tuning}
Due to the limited amount of data available, we must exercise caution when modifying the parameters of our pretrained model. We find that updating only a few parameters (i.e., layers) is effective. As discussed in \citet{kustowski2022suppressing}, updating a single layer of the neural network, rather than all the parameters of the model, is essential to avoid overfitting. 

To fine-tune our model $f_{\theta^s}$ on the experimental dataset $\mathcal{R}$, we employ a leave-one-out cross-validation strategy, given the small size of our dataset $D^t$ which consists of $N=10$ samples. In this process, we use $9$ samples for training and $1$ sample for testing. During training, we compute a validation error by performing another round of leave-one-out validation, where we fine-tune a model on $8$ of the $9$ training points, and then evaluate on the held-out point. 

As defined above, we specify a fully train model to be
 $\theta^* = [\beta^s_{\mathrm{fixed}},\beta^*_{\mathrm{trainable}}]$

\begin{equation}
\beta^*_{\mathrm{trainable}} = \left\{
\begin{array}{ll}
\beta_0 = \beta^s_{\mathrm{trainable}} & \mathrm{Initialize} \\
\beta_{j+1} = \beta_j - \delta \nabla L_{pred}\;\ref{eqn:Lpred}, & j = 0, 1, \ldots, E-1
\end{array}
\right.
\end{equation}

Where $\delta$ is the learning rate to update just the trainable parameters $\beta_j$ and we use the $L_{pred}$ loss function (\ref{eqn:Lpred}) and set $\gamma_o = 0$ or $\gamma_i = 0$. This zeroing out to focus on one modality is employed to avoid overfitting on scalars at the expense of images or vice versa, degrading the model's overall performance. By focusing on each modality individually, we ensure that the model can learn and capture the unique characteristics of each data type without being negatively influenced by the other. We investigated fine-tuning our model on both the images and scalars simultaneously. However, we found that this approach resulted in inferior performance compared to the separate training on images and scalars.

Finally, we repeat this process for all $9$ training points, then average the error of the held-out validation points $V = \frac{1}{N}\sum_{j=1}^N V_j$. This approach allows us to systematically evaluate the model's performance across different experimental data splits while making the best use of the limited available data.

\paragraph{Hyper-parameter Grid Search}
During the fine-tuning process, we perform a grid search over a range of hyper-parameters. The aim of the grid search is to identify the optimal combination of hyper-parameters that yield the best performance on the validation set. Some of the hyper-parameters explored during the grid search include the learning rate, the number of fine-tuning epochs, and determining which layer to tune. By exhaustively searching the grid over the hyper-parameter space, we ensure an optimal model can be selected for a given training set.

Finally, due to the few-shot nature of our data, we fine-tune our model on both the training and validation data using the selected hyper-parameters. After the fine-tuning process is complete, we evaluate the performance of our model on the held-out test set. This provides us with an estimate of the model's generalization capability when applied to unseen experimental data.

\paragraph{Early Stopping Post-Hoc Correction}
As we often stop fine-tuning a model before it finds a local minimum of the loss function, we found these models to consistently underfit the training data. To counteract this deficiency in model fit, we propose a method that manually adjusts the bias and variance of the predictions in accordance with the training set. The primary idea behind this approach is to strike an optimal balance between overfitting (halting training when the prediction loss ceases to decrease) and underfitting (insufficient updates to the model weights to account for bias). We suggest a straightforward solution that involves manually modifying the model's final predictions using new variance and bias parameters.


We compute the average error from the training data for each predicted scalar $b^k = \frac{1}{N} \sum_{j=1}^n f_{\theta}^{t}(x_j)^k - y_j^k$, where $f(x_j)^k$ represents the $k$-th scalar output of the finetuned model $f_{\theta}^{t}$, and adjust the final validation set predictions to account for this average error over the $n$ training points: 

\begin{equation}
\hat{y}^k = f_{\theta}^{t}(x)^k - b^k
\end{equation}

A similar approach is applied to the variance of the predictions. Let the average for scalar $k$ be $\mu_k = \frac{1}{N} \sum_j^n f_{surr}(x_j)^k$ and the variance be $\sigma(y^k) = var({y_0^k,y_1^k, ... ,y_n^k })$:

\begin{equation}
\hat{y}^k = \mu^k + ((f(x)^k - \mu^k) * \frac{\sigma(y^k)}{\sigma(\hat{y}^k)} )
\end{equation}

\subsection{Implementation and Dataset Details}
In our implementation, we employ a variant of the Masked Autoencoder (MAE) that closely aligns with the popular architecture proposed by \citet{he2022masked}, albeit with modifications to suit our multi-modal dataset and computational constraints. Specifically, our MAE model is characterized by a reduced number of decoder blocks (6) and smaller embedding sizes, with $512$ dimensions for the encoder and $256$ dimensions for the decoder. Our decision to opt for a smaller model was chosen based on empirical evidence from preliminary experiments and is also informed by the broader observation in the field that, beyond a certain point, larger embedding sizes do not equate to significant performance improvements, particularly for datasets of moderate size and dimensionality. The hardware used for training comprised a single NVIDIA V100 GPU, with hyper-parameter tuning and experimentation facilitated by parallelization across a cluster of 64 V100 GPUs. 

The Adam optimizer is employed with a cosine annealed learning rate starting at $10^{-3}$ (which gradually decreases to 0). 
The best model is selected based on the pre-training simulation test set average error rate (optimized over different hyper-parameters: $\gamma_o$, epochs, and learning rates). For each leave-one-out test set experiment, we select the best smoothed validation score as described above.

For the $\real$ dataset, the large simulation database, $\mathcal{S}$, was created using the two-dimensional radiation hydrodynamic code HYDRA~\cite{marinak2001three}. These simulations serve as an extensive sampling of the design space, permitting more robust predictive modeling.

Our second dataset, $\synth$, is generated synthetically. It is designed to create a representative set of ICF experiments by employing an uncalibrated surrogate model. Instead of running new HYDRA simulations, which would be computationally expensive and time-consuming, \citet{kustowski2022suppressing} utilized their uncalibrated surrogate model to make predictions. This approach enabled them to create two lower-dimensional and physically inconsistent datasets for transfer learning, which are nearly equivalent to running a new set of simulations.
To create the synthetic datasets, they fixed four of the nine input parameters and sampled the remaining five input parameters randomly within their original ranges. They used their uncalibrated surrogate model to predict the outputs and, then, perturbed the values of the asymmetry and preheat parameters to create 1,000 "experiments". 

The pretraining simulation dataset comprises of $90000$ training samples and $2000$ test samples. Images are X-rays of $60x60$ pixels and are self normalized, with each image's pixels divided by its own mean, as each image may span differing orders of magnitude.
The experimental dataset consists of 10 samples, which are divided using leave-one-out for training.
The synthetic dataset includes $1000$ samples. To stay consistent with the experimental dataset, we only fine-tune with a few samples (7,9, or 50); and we report the average error over the other held-out points.



\subsection{A novel Graph-Based approach for Robust Hyper-parameter Selection}
\label{sec:hpo}
Given a set of candidate hyper-parameter configurations, we construct a graph $G = (\mathcal{V}, \mathcal{E})$, where each node $v_i \in \mathcal{V}$ represents a unique hyper-parameter configuration $\lambda_i$, and an edge $(v_i, v_j) \in \mathcal{E}$ exists if the corresponding configurations differ in exactly one hyper-parameter setting by a single step in that hyper-parameter.
For example, an edge would exist between two hyper-parameter configures if they only differ in the learning rate by one step (e.g., $10^{-3}$ or $10^{-4}$). There would not be an edge between a two-step size difference, such as $10^{-3}$ and $10^{-5}$. In addition, there would not be an edge if two hyper-parameters were changed; for example, if the learning rate and the epochs to train were different between two fine-tuning runs, then no edge would be between the two nodes corresponding to these two hyper-parameter configurations. This graph helps us understand the local structure of the hyper-parameter space and how small changes in the configurations are related.

The hyper-parameters we use are as follows:
\begin{enumerate}
    \item Transformer decoder block to train (1-7)
    \item Epochs to train (5,10,20,30,40,50,75,100,200,300,400,500)
    \item Learning rate ($10^{-3},10^{-4},10^{-5}$)
    \item Fine-tuning loss function (L1,L2, Huber)
    \item Use post-hoc correction (bias and/or variance)
    \item Pretraining $\alpha$ ($0.02$ or $1.0$)
\end{enumerate}

In our study, the determination of hyper-parameter grid points for exhaustive scans was initially guided by a trial-and-error approach, resulting in a comprehensive exploration across 6,048 hyper-parameter configurations for each experiment. Recognizing the potential inefficiencies of this method, we propose a more systematic approach for future work and practitioners aiming to optimize the hyper-parameter selection process. Specifically, employing Bayesian optimization offers a promising starting point for identifying promising regions within the hyper-parameter space. This probabilistic model-based approach can effectively suggest initial values that are likely to yield improved performance metrics. Following the identification of these regions, an exponential or binary search strategy could be implemented to refine the grid resolution. 

Validation error rates are separately computed for both images and scalars. The error for an image is simply the MSE averaged over all the pixels, and the error for the scalars is the MSE averaged over the ten target scalars. The validation error is the average error from doing a leave-one-out cross validation on the training set.
We separate the validation error rates between images and scalars to keep in line with our separate training process described above. For the sake of clarity, the following description only considers a single validation score (e.g., the image MSE).
We assign node values based on the validation error rates, denoted by $\mathbf{V} = \{V_1, \ldots, V_n\}$, where $V_j = \frac{1}{N}\sum_{j=1}^N V_j$ corresponds to the validation error rate for the hyper-parameter configuration $\lambda_j$ averaged over the $N$ leave-one-out experiments for a given training set. The minimum validation error rate configuration is defined as:
\begin{equation}
\val= \arg\min_{i} {V}_i
\end{equation}

Next, to exploit the graph structure for hyper-parameter optimization, we perform a simple smoothing on the graph $G$. This process updates the node values by considering both the original validation error rate and the average value of neighboring nodes.

Let $\mathbf{A}$ be the adjacency matrix of the graph $G$, and $\mathcal{N}(i)$ denote the set of neighbors of node $i$. We define the smoothed node value $\tilde{V}_i$ as follows:

\begin{equation}
\tilde{V}_i = \frac{1}{2} V_i + \frac{1}{2} \frac{\sum_{j \in \mathcal{N}(i)} \mathbf{A}_{ij} V_j}{|\mathcal{N}(i)|}
\end{equation}

where $A_{ij}$ denotes the element of the adjacency matrix at position $(i, j)$. The first term in the equation represents half of the original validation error value, while the second term represents half of the average neighbor value.

After applying the smoothing, we select the hyper-parameter configuration corresponding to the node with the lowest smoothed value:

\begin{equation}
\graph = \arg\min_{i} \tilde{V}_i
\end{equation}

The selected configuration $\graph$ represents an optimal choice that balances the original validation error rates and the information propagated from neighboring nodes. This graph-based approach is particularly beneficial in the context of few-shot learning, where the limited number of examples can lead to noisy estimates of model performance. By exploiting the structure of the hyper-parameter space, our method effectively identifies optimal hyper-parameter configurations and consistently improves the overall performance for our few-shot scenario. Our proposed design is based on the premise that we have a comprehensive grid search over the hyper-parameters of interest. This choice of exploration strategy lends itself naturally to the construction of the graph, where each node represents a unique hyper-parameter configuration and edges connect nodes that differ in exactly one dimension by a single parameter step. This approach results in a well-defined neighborhood structure that captures the local similarities between configurations.
However, it is important to note that more complex neighboring strategies could be employed when dealing with more sophisticated hyper-parameter sweep settings, such as random search or Bayesian Optimization \citep{snoek2012practical}. In such cases, alternative techniques for defining the connectivity between nodes might be required to capture the relationships between different configurations.

In our analysis, we focus on using the fewest neighbors possible in order to balance the exploitation of the graph structure and the preservation of the original validation error rates. This choice is motivated by the desire to avoid over-smoothing, which can lead to suboptimal hyper-parameter configurations. 


\section{Acknowledgements}
This work was performed under the auspices of the U.S. Department of Energy by Lawrence Livermore National Laboratory under Contract DE-AC52-07NA27344. The work is supported by Laboratory Directed Research and Development Program (LDRD) 22-ERD-006, and supported by DOE FES Measurements Innovations grant SCW1720. IM Release number LLNL-JRNL-848991.

\bibliographystyle{IEEEtranN}
\bibliography{001-bibliography}

\begin{thebibliography}{46}
\providecommand{\natexlab}[1]{#1}
\providecommand{\url}[1]{#1}
\csname url@samestyle\endcsname
\providecommand{\newblock}{\relax}
\providecommand{\bibinfo}[2]{#2}
\providecommand{\BIBentrySTDinterwordspacing}{\spaceskip=0pt\relax}
\providecommand{\BIBentryALTinterwordstretchfactor}{4}
\providecommand{\BIBentryALTinterwordspacing}{\spaceskip=\fontdimen2\font plus
\BIBentryALTinterwordstretchfactor\fontdimen3\font minus
  \fontdimen4\font\relax}
\providecommand{\BIBforeignlanguage}[2]{{%
\expandafter\ifx\csname l@#1\endcsname\relax
\typeout{** WARNING: IEEEtranN.bst: No hyphenation pattern has been}%
\typeout{** loaded for the language `#1'. Using the pattern for}%
\typeout{** the default language instead.}%
\else
\language=\csname l@#1\endcsname
\fi
#2}}
\providecommand{\BIBdecl}{\relax}
\BIBdecl

\bibitem[Hatfield et~al.(2021)Hatfield, Gaffney, Anderson, Ali, Antonelli,
  Ba{\c{s}}e{\u{g}}mez~du Pree, Citrin, Fajardo, Knapp, Kettle,
  et~al.]{hatfield2021data}
P.~W. Hatfield, J.~A. Gaffney, G.~J. Anderson, S.~Ali, L.~Antonelli,
  S.~Ba{\c{s}}e{\u{g}}mez~du Pree, J.~Citrin, M.~Fajardo, P.~Knapp, B.~Kettle
  \emph{et~al.}, ``The data-driven future of high-energy-density physics,''
  \emph{Nature}, vol. 593, no. 7859, pp. 351--361, 2021.

\bibitem[Nora et~al.(2017)Nora, Peterson, Spears, Field, and
  Brandon]{nora2017ensemble}
R.~Nora, J.~L. Peterson, B.~K. Spears, J.~E. Field, and S.~Brandon, ``Ensemble
  simulations of inertial confinement fusion implosions,'' \emph{Statistical
  Analysis and Data Mining: The ASA Data Science Journal}, vol.~10, no.~4, pp.
  230--237, 2017.

\bibitem[Humbird et~al.(2019)Humbird, Peterson, Spears, and
  McClarren]{humbird2019transfer}
K.~D. Humbird, J.~L. Peterson, B.~Spears, and R.~G. McClarren, ``Transfer
  learning to model inertial confinement fusion experiments,'' \emph{IEEE
  Transactions on Plasma Science}, vol.~48, no.~1, pp. 61--70, 2019.

\bibitem[Kustowski et~al.(2019)Kustowski, Gaffney, Spears, Anderson,
  Thiagarajan, and Anirudh]{kustowski2019transfer}
B.~Kustowski, J.~A. Gaffney, B.~K. Spears, G.~J. Anderson, J.~J. Thiagarajan,
  and R.~Anirudh, ``Transfer learning as a tool for reducing simulation bias:
  application to inertial confinement fusion,'' \emph{IEEE Transactions on
  Plasma Science}, vol.~48, no.~1, pp. 46--53, 2019.

\bibitem[Kustowski et~al.(2022)Kustowski, Gaffney, Spears, Anderson, Anirudh,
  Bremer, Thiagarajan, Kruse, and Nora]{kustowski2022suppressing}
B.~Kustowski, J.~A. Gaffney, B.~K. Spears, G.~J. Anderson, R.~Anirudh, P.-T.
  Bremer, J.~J. Thiagarajan, M.~K. Kruse, and R.~C. Nora, ``Suppressing
  simulation bias in multi-modal data using transfer learning,'' \emph{Machine
  Learning: Science and Technology}, vol.~3, no.~1, p. 015035, 2022.

\bibitem[Schmidt and Lipson(2009)]{schmidt2009distilling}
M.~Schmidt and H.~Lipson, ``Distilling free-form natural laws from experimental
  data,'' \emph{science}, vol. 324, no. 5923, pp. 81--85, 2009.

\bibitem[Pan and Yang(2009)]{pan2009survey}
S.~J. Pan and Q.~Yang, ``A survey on transfer learning,'' \emph{IEEE
  Transactions on knowledge and data engineering}, vol.~22, no.~10, pp.
  1345--1359, 2009.

\bibitem[Trivedi et~al.(2023)Trivedi, Koutra, and
  Thiagarajan]{trivedi2023closer}
P.~Trivedi, D.~Koutra, and J.~J. Thiagarajan, ``A closer look at model
  adaptation using feature distortion and simplicity bias,'' \emph{arXiv
  preprint arXiv:2303.13500}, 2023.

\bibitem[Betti and Hurricane(2016)]{betti2016inertial}
R.~Betti and O.~Hurricane, ``Inertial-confinement fusion with lasers,''
  \emph{Nature Physics}, vol.~12, no.~5, pp. 435--448, 2016.

\bibitem[Vaswani et~al.(2017)Vaswani, Shazeer, Parmar, Uszkoreit, Jones, Gomez,
  Kaiser, and Polosukhin]{vaswani2017attention}
A.~Vaswani, N.~Shazeer, N.~Parmar, J.~Uszkoreit, L.~Jones, A.~N. Gomez,
  {\L}.~Kaiser, and I.~Polosukhin, ``Attention is all you need,''
  \emph{Advances in neural information processing systems}, vol.~30, 2017.

\bibitem[Devlin et~al.(2018)Devlin, Chang, Lee, and Toutanova]{devlin2018bert}
J.~Devlin, M.-W. Chang, K.~Lee, and K.~Toutanova, ``Bert: Pre-training of deep
  bidirectional transformers for language understanding,'' \emph{arXiv preprint
  arXiv:1810.04805}, 2018.

\bibitem[Radford et~al.(2018)Radford, Narasimhan, Salimans, and
  Sutskever]{radford2018improving}
A.~Radford, K.~Narasimhan, T.~Salimans, and I.~Sutskever, ``Improving language
  understanding with unsupervised learning,'' 2018.

\bibitem[Radford et~al.(2019)Radford, Wu, Child, Luan, Amodei, and
  Sutskever]{radford2019language}
A.~Radford, J.~Wu, R.~Child, D.~Luan, D.~Amodei, and I.~Sutskever, ``Language
  models are unsupervised multitask learners,'' \emph{OpenAI blog}, vol.~1,
  no.~8, p.~9, 2019.

\bibitem[Brown et~al.(2020)Brown, Mann, Ryder, Subbiah, Kaplan, Dhariwal,
  Neelakantan, Shyam, Sastry, Askell, Agarwal, Herbert-Voss, Krueger, Henighan,
  Child, Ramesh, Ziegler, Wu, Winter, Hesse, Chen, Sigler, Litwin, Gray, Chess,
  Clark, Berner, McCandlish, Radford, Sutskever, and Amodei]{brown2020language}
T.~Brown, B.~Mann, N.~Ryder, M.~Subbiah, J.~D. Kaplan, P.~Dhariwal,
  A.~Neelakantan, P.~Shyam, G.~Sastry, A.~Askell, S.~Agarwal, A.~Herbert-Voss,
  G.~Krueger, T.~Henighan, R.~Child, A.~Ramesh, D.~Ziegler, J.~Wu, C.~Winter,
  C.~Hesse, M.~Chen, E.~Sigler, M.~Litwin, S.~Gray, B.~Chess, J.~Clark,
  C.~Berner, S.~McCandlish, A.~Radford, I.~Sutskever, and D.~Amodei, ``Language
  models are few-shot learners,'' \emph{Advances in neural information
  processing systems}, vol.~33, pp. 1877--1901, 2020.

\bibitem[Bubeck et~al.(2023)Bubeck, Chandrasekaran, Eldan, Gehrke, Horvitz,
  Kamar, Lee, Lee, Li, Lundberg, Nori, Palangi, Ribeiro, and
  Zhang]{bubeck2023sparks}
S.~Bubeck, V.~Chandrasekaran, R.~Eldan, J.~Gehrke, E.~Horvitz, E.~Kamar,
  P.~Lee, Y.~T. Lee, Y.~Li, S.~Lundberg, H.~Nori, H.~Palangi, M.~T. Ribeiro,
  and Y.~Zhang, ``Sparks of artificial general intelligence: Early experiments
  with gpt-4,'' \emph{arXiv preprint arXiv:2303.12712}, 2023.

\bibitem[Dosovitskiy et~al.(2021)Dosovitskiy, Beyer, Kolesnikov, Weissenborn,
  Zhai, Unterthiner, Dehghani, Minderer, Heigold, Gelly, Uszkoreit, and
  Houlsby]{dosovitskiy2021an}
\BIBentryALTinterwordspacing
A.~Dosovitskiy, L.~Beyer, A.~Kolesnikov, D.~Weissenborn, X.~Zhai,
  T.~Unterthiner, M.~Dehghani, M.~Minderer, G.~Heigold, S.~Gelly, J.~Uszkoreit,
  and N.~Houlsby, ``An image is worth 16x16 words: Transformers for image
  recognition at scale,'' in \emph{International Conference on Learning
  Representations}, 2021. [Online]. Available:
  \url{https://openreview.net/forum?id=YicbFdNTTy}
\BIBentrySTDinterwordspacing

\bibitem[Zhai et~al.(2022)Zhai, Kolesnikov, Houlsby, and
  Beyer]{zhai2022scaling}
X.~Zhai, A.~Kolesnikov, N.~Houlsby, and L.~Beyer, ``Scaling vision
  transformers,'' in \emph{Proceedings of the IEEE/CVF Conference on Computer
  Vision and Pattern Recognition}, 2022, pp. 12\,104--12\,113.

\bibitem[Khan et~al.(2022)Khan, Naseer, Hayat, Zamir, Khan, and
  Shah]{khan2022transformers}
S.~Khan, M.~Naseer, M.~Hayat, S.~W. Zamir, F.~S. Khan, and M.~Shah,
  ``Transformers in vision: A survey,'' \emph{ACM computing surveys (CSUR)},
  vol.~54, no. 10s, pp. 1--41, 2022.

\bibitem[Fang et~al.(2021)Fang, Liao, Wang, Fang, Qi, Wu, Niu, and
  Liu]{fang2021you}
Y.~Fang, B.~Liao, X.~Wang, J.~Fang, J.~Qi, R.~Wu, J.~Niu, and W.~Liu, ``You
  only look at one sequence: Rethinking transformer in vision through object
  detection,'' \emph{Advances in Neural Information Processing Systems},
  vol.~34, pp. 26\,183--26\,197, 2021.

\bibitem[Dhariwal et~al.(2020)Dhariwal, Jun, Payne, Kim, Radford, and
  Sutskever]{dhariwal2020jukebox}
P.~Dhariwal, H.~Jun, C.~Payne, J.~W. Kim, A.~Radford, and I.~Sutskever,
  ``Jukebox: A generative model for music,'' \emph{arXiv preprint
  arXiv:2005.00341}, 2020.

\bibitem[Kreuk et~al.(2022)Kreuk, Synnaeve, Polyak, Singer, D{\'e}fossez,
  Copet, Parikh, Taigman, and Adi]{kreuk2022audiogen}
F.~Kreuk, G.~Synnaeve, A.~Polyak, U.~Singer, A.~D{\'e}fossez, J.~Copet,
  D.~Parikh, Y.~Taigman, and Y.~Adi, ``Audiogen: Textually guided audio
  generation,'' \emph{arXiv preprint arXiv:2209.15352}, 2022.

\bibitem[Borsos et~al.(2023)Borsos, Marinier, Vincent, Kharitonov, Pietquin,
  Sharifi, Roblek, Teboul, Grangier, Tagliasacchi, et~al.]{borsos2023audiolm}
Z.~Borsos, R.~Marinier, D.~Vincent, E.~Kharitonov, O.~Pietquin, M.~Sharifi,
  D.~Roblek, O.~Teboul, D.~Grangier, M.~Tagliasacchi \emph{et~al.}, ``Audiolm:
  a language modeling approach to audio generation,'' \emph{IEEE/ACM
  Transactions on Audio, Speech, and Language Processing}, 2023.

\bibitem[Schwaller et~al.(2019)Schwaller, Laino, Gaudin, Bolgar, Hunter, Bekas,
  and Lee]{schwaller2019molecular}
P.~Schwaller, T.~Laino, T.~Gaudin, P.~Bolgar, C.~A. Hunter, C.~Bekas, and A.~A.
  Lee, ``Molecular transformer: a model for uncertainty-calibrated chemical
  reaction prediction,'' \emph{ACS central science}, vol.~5, no.~9, pp.
  1572--1583, 2019.

\bibitem[Schwaller et~al.(2021{\natexlab{a}})Schwaller, Probst, Vaucher, Nair,
  Kreutter, Laino, and Reymond]{schwaller2021mapping}
P.~Schwaller, D.~Probst, A.~C. Vaucher, V.~H. Nair, D.~Kreutter, T.~Laino, and
  J.-L. Reymond, ``Mapping the space of chemical reactions using
  attention-based neural networks,'' \emph{Nature machine intelligence},
  vol.~3, no.~2, pp. 144--152, 2021.

\bibitem[Schwaller et~al.(2021{\natexlab{b}})Schwaller, Hoover, Reymond,
  Strobelt, and Laino]{schwaller2021extraction}
P.~Schwaller, B.~Hoover, J.-L. Reymond, H.~Strobelt, and T.~Laino, ``Extraction
  of organic chemistry grammar from unsupervised learning of chemical
  reactions,'' \emph{Science Advances}, vol.~7, no.~15, p. eabe4166, 2021.

\bibitem[Born and Manica(2023)]{born2023regression}
J.~Born and M.~Manica, ``Regression transformer enables concurrent sequence
  regression and generation for molecular language modelling,'' \emph{Nature
  Machine Intelligence}, vol.~5, no.~4, pp. 432--444, 2023.

\bibitem[Rives et~al.(2021)Rives, Meier, Sercu, Goyal, Lin, Liu, Guo, Ott,
  Zitnick, Ma, et~al.]{rives2021biological}
A.~Rives, J.~Meier, T.~Sercu, S.~Goyal, Z.~Lin, J.~Liu, D.~Guo, M.~Ott, C.~L.
  Zitnick, J.~Ma \emph{et~al.}, ``Biological structure and function emerge from
  scaling unsupervised learning to 250 million protein sequences,''
  \emph{Proceedings of the National Academy of Sciences}, vol. 118, no.~15, p.
  e2016239118, 2021.

\bibitem[Jumper et~al.(2021)Jumper, Evans, Pritzel, Green, Figurnov,
  Ronneberger, Tunyasuvunakool, Bates, {\v{Z}}{\'\i}dek, Potapenko,
  et~al.]{jumper2021highly}
J.~Jumper, R.~Evans, A.~Pritzel, T.~Green, M.~Figurnov, O.~Ronneberger,
  K.~Tunyasuvunakool, R.~Bates, A.~{\v{Z}}{\'\i}dek, A.~Potapenko
  \emph{et~al.}, ``Highly accurate protein structure prediction with
  alphafold,'' \emph{Nature}, vol. 596, no. 7873, pp. 583--589, 2021.

\bibitem[He et~al.(2022)He, Chen, Xie, Li, Doll{\'a}r, and
  Girshick]{he2022masked}
K.~He, X.~Chen, S.~Xie, Y.~Li, P.~Doll{\'a}r, and R.~Girshick, ``Masked
  autoencoders are scalable vision learners,'' in \emph{Proceedings of the
  IEEE/CVF Conference on Computer Vision and Pattern Recognition}, 2022, pp.
  16\,000--16\,009.

\bibitem[Atzeni and Meyer-ter Vehn(2004)]{atzeni2004physics}
\BIBentryALTinterwordspacing
S.~Atzeni and J.~Meyer-ter Vehn, \emph{The Physics of Inertial Fusion: Beam
  Plasma Interaction, Hydrodynamics, Hot Dense Matter}, ser. International
  Series of Monographs on Physics.\hskip 1em plus 0.5em minus 0.4em\relax OUP
  Oxford, 2004. [Online]. Available:
  \url{https://books.google.com/books?id=BJcy\_p5pUBsC}
\BIBentrySTDinterwordspacing

\bibitem[Casey et~al.(2018)Casey, Thomas, Baker, Spears, Hohenberger, Khan,
  Nora, Weber, Woods, Hurricane, et~al.]{casey2018high}
D.~Casey, C.~Thomas, K.~Baker, B.~Spears, M.~Hohenberger, S.~Khan, R.~Nora,
  C.~Weber, D.~Woods, O.~Hurricane \emph{et~al.}, ``The high velocity, high
  adiabat,“bigfoot” campaign and tests of indirect-drive implosion
  scaling,'' \emph{Physics of Plasmas}, vol.~25, no.~5, p. 056308, 2018.

\bibitem[Anirudh et~al.(2020)Anirudh, Thiagarajan, Bremer, and
  Spears]{anirudh2020improved}
R.~Anirudh, J.~J. Thiagarajan, P.-T. Bremer, and B.~K. Spears, ``Improved
  surrogates in inertial confinement fusion with manifold and cycle
  consistencies,'' \emph{Proceedings of the National Academy of Sciences}, vol.
  117, no.~18, pp. 9741--9746, 2020.

\bibitem[Kornblith et~al.(2019)Kornblith, Norouzi, Lee, and
  Hinton]{kornblith2019similarity}
S.~Kornblith, M.~Norouzi, H.~Lee, and G.~Hinton, ``Similarity of neural network
  representations revisited,'' in \emph{International conference on machine
  learning}.\hskip 1em plus 0.5em minus 0.4em\relax PMLR, 2019, pp. 3519--3529.

\bibitem[Davari et~al.(2023)Davari, Horoi, Natik, Lajoie, Wolf, and
  Belilovsky]{davari2023reliability}
\BIBentryALTinterwordspacing
M.~Davari, S.~Horoi, A.~Natik, G.~Lajoie, G.~Wolf, and E.~Belilovsky,
  ``Reliability of {CKA} as a similarity measure in deep learning,'' in
  \emph{The Eleventh International Conference on Learning Representations},
  2023. [Online]. Available: \url{https://openreview.net/forum?id=8HRvyxc606}
\BIBentrySTDinterwordspacing

\bibitem[Franceschi et~al.(2018)Franceschi, Frasconi, Salzo, Grazzi, and
  Pontil]{franceschi2018bilevel}
L.~Franceschi, P.~Frasconi, S.~Salzo, R.~Grazzi, and M.~Pontil, ``Bilevel
  programming for hyperparameter optimization and meta-learning,'' in
  \emph{International Conference on Machine Learning}.\hskip 1em plus 0.5em
  minus 0.4em\relax PMLR, 2018, pp. 1568--1577.

\bibitem[Mazumder et~al.(2021)Mazumder, Singh, and
  Namboodiri]{mazumder2021rnnp}
P.~Mazumder, P.~Singh, and V.~P. Namboodiri, ``Rnnp: A robust few-shot learning
  approach,'' in \emph{Proceedings of the IEEE/CVF Winter Conference on
  Applications of Computer Vision}, 2021, pp. 2664--2673.

\bibitem[Van~Rijn and Hutter(2018)]{van2018hyperparameter}
J.~N. Van~Rijn and F.~Hutter, ``Hyperparameter importance across datasets,'' in
  \emph{Proceedings of the 24th ACM SIGKDD International Conference on
  Knowledge Discovery \& Data Mining}, 2018, pp. 2367--2376.

\bibitem[Liang et~al.(2022)Liang, Rangrej, Petrovic, and
  Hassner]{Liang_2022_CVPR}
K.~J. Liang, S.~B. Rangrej, V.~Petrovic, and T.~Hassner, ``Few-shot learning
  with noisy labels,'' in \emph{Proceedings of the IEEE/CVF Conference on
  Computer Vision and Pattern Recognition (CVPR)}, June 2022, pp. 9089--9098.

\bibitem[Muniraju et~al.(2020)Muniraju, Kailkhura, Thiagarajan, Bremer,
  Tepedelenlioglu, and Spanias]{muniraju2020coverage}
G.~Muniraju, B.~Kailkhura, J.~J. Thiagarajan, P.-T. Bremer, C.~Tepedelenlioglu,
  and A.~Spanias, ``Coverage-based designs improve sample mining and
  hyperparameter optimization,'' \emph{IEEE Transactions on Neural Networks and
  Learning Systems}, vol.~32, no.~3, pp. 1241--1253, 2020.

\bibitem[Booker et~al.(1999)Booker, Dennis, Frank, Serafini, Torczon, and
  Trosset]{booker1999rigorous}
A.~J. Booker, J.~E. Dennis, P.~D. Frank, D.~B. Serafini, V.~Torczon, and M.~W.
  Trosset, ``A rigorous framework for optimization of expensive functions by
  surrogates,'' \emph{Structural optimization}, vol.~17, pp. 1--13, 1999.

\bibitem[Kennedy and O'Hagan(2001)]{kennedy2001bayesian}
M.~C. Kennedy and A.~O'Hagan, ``Bayesian calibration of computer models,''
  \emph{Journal of the Royal Statistical Society: Series B (Statistical
  Methodology)}, vol.~63, no.~3, pp. 425--464, 2001.

\bibitem[Radford et~al.(2021)Radford, Kim, Hallacy, Ramesh, Goh, Agarwal,
  Sastry, Askell, Mishkin, Clark, et~al.]{radford2021learning}
A.~Radford, J.~W. Kim, C.~Hallacy, A.~Ramesh, G.~Goh, S.~Agarwal, G.~Sastry,
  A.~Askell, P.~Mishkin, J.~Clark \emph{et~al.}, ``Learning transferable visual
  models from natural language supervision,'' in \emph{International conference
  on machine learning}.\hskip 1em plus 0.5em minus 0.4em\relax PMLR, 2021, pp.
  8748--8763.

\bibitem[Li et~al.(2019)Li, Liu, Liu, Zhao, and Liu]{li2019neural}
N.~Li, S.~Liu, Y.~Liu, S.~Zhao, and M.~Liu, ``Neural speech synthesis with
  transformer network,'' in \emph{Proceedings of the AAAI conference on
  artificial intelligence}, vol.~33, no.~01, 2019, pp. 6706--6713.

\bibitem[Ba et~al.(2016)Ba, Kiros, and Hinton]{ba2016layer}
J.~L. Ba, J.~R. Kiros, and G.~E. Hinton, ``Layer normalization,'' \emph{arXiv
  preprint arXiv:1607.06450}, 2016.

\bibitem[Marinak et~al.(2001)Marinak, Kerbel, Gentile, Jones, Munro, Pollaine,
  Dittrich, and Haan]{marinak2001three}
M.~M. Marinak, G.~Kerbel, N.~Gentile, O.~Jones, D.~Munro, S.~Pollaine,
  T.~Dittrich, and S.~Haan, ``Three-dimensional hydra simulations of national
  ignition facility targets,'' \emph{Physics of Plasmas}, vol.~8, no.~5, pp.
  2275--2280, 2001.

\bibitem[Snoek et~al.(2012)Snoek, Larochelle, and Adams]{snoek2012practical}
J.~Snoek, H.~Larochelle, and R.~P. Adams, ``Practical bayesian optimization of
  machine learning algorithms,'' \emph{Advances in neural information
  processing systems}, vol.~25, 2012.

\end{thebibliography}


\end{document}


\title[Supplement: Transformer-Powered Surrogate Design]{Supplement for Transformer-Powered Surrogates Close the ICF Simulation-Experiment Gap with Extremely Limited Data}

\author{Matthew L. Olson, Shusen Liu, Jayaraman J. Thiagarajan,  Bogdan Kustowski, Weng-Keen Wong, Rushil Anirudh}

\address{Lawrence Livermore National Laboratory}
\ead{olson60@llnl.gov}

\section{Detailed tables}
In table \ref{tbl:loo_val_vs_graph}, we look at a detailed version of experiments comparing \mval versus \mgraph.
In table \ref{tbl:leave3out} we show the detailed results of leave-3-out.
In tables \ref{tbl:results_leave1_out_real} and \ref{tbl:results_leave1_out_sim} we show detailed results for real and synth datasets respectively.

\begin{table}[htbp]
\centering
\begin{tabular}{ l c c c c}
Scalar Name            &  $\val (\real)$ & $\graph (\real)$   & $\val (\synth)$ & $\graph (\synth)$ \\ \hline
Neutron bang time      & 0.048{\scriptsize$\pm 0.068$ }& 0.037{\scriptsize$\pm 0.041$ } & 0.636{\scriptsize$\pm 0.230$ }& 0.664{\scriptsize$\pm 0.276$ }                        \\ 
X-ray bang time        & 0.036{\scriptsize$\pm 0.053$ }& 0.029{\scriptsize$\pm 0.027$ } & 0.647{\scriptsize$\pm 0.201$ }& 0.679{\scriptsize$\pm 0.259$ }                        \\ 
Downscattered ratio    & 0.705{\scriptsize$\pm 0.741$ }& 0.550{\scriptsize$\pm 0.541$ } & 5.875{\scriptsize$\pm 3.577$ }& 4.495{\scriptsize$\pm 1.753$ }                        \\ 
Temperature            & 0.175{\scriptsize$\pm 0.178$ }& 0.152{\scriptsize$\pm 0.136$ } & 2.815{\scriptsize$\pm 0.966$ }& 2.893{\scriptsize$\pm 0.722$ }                        \\ 
Hot spot radius        & 0.110{\scriptsize$\pm 0.139$ }& 0.116{\scriptsize$\pm 0.125$ } & 7.187{\scriptsize$\pm 1.670$ }& 6.788{\scriptsize$\pm 1.681$ }                        \\ 
Velocity               & 0.331{\scriptsize$\pm 0.387$ }& 0.212{\scriptsize$\pm 0.218$ } & 7.657{\scriptsize$\pm 3.885$ }& 6.970{\scriptsize$\pm 2.758$ }                        \\ 
X-ray emission         & 1.242{\scriptsize$\pm 1.818$ }& 0.745{\scriptsize$\pm 1.099$ } & 4.191{\scriptsize$\pm 1.943$ }& 4.516{\scriptsize$\pm 1.952$ }                        \\ 
Neutron yield          & 0.145{\scriptsize$\pm 0.138$ }& 0.035{\scriptsize$\pm 0.038$ } & 3.814{\scriptsize$\pm 1.699$ }& 4.355{\scriptsize$\pm 1.764$ }                        \\ 
Neutron burn width     & 0.269{\scriptsize$\pm 0.278$ }& 0.320{\scriptsize$\pm 0.424$ } & 10.066{\scriptsize$\pm 3.162$ }& 8.851{\scriptsize$\pm 4.546$ }                       \\ 
X-ray burn width       & 2.783{\scriptsize$\pm 4.537$ }& 2.728{\scriptsize$\pm 3.902$ } & 12.769{\scriptsize$\pm 4.548$ }& 11.342{\scriptsize$\pm 5.413$ }                      \\
Average & 0.584{\scriptsize$\pm 0.813$ }& 0.492{\scriptsize$\pm 0.779$ }  & 5.566{\scriptsize$\pm 3.745$ }& 5.155{\scriptsize$\pm 3.236$ }                        \\ \hline
Images                 & 0.158{\scriptsize$\pm 0.102$ }& 0.154{\scriptsize$\pm 0.095$ }  & 0.031{\scriptsize$\pm 0.009$ }& 0.030{\scriptsize$\pm 0.009$ }                        \\ \hline
\end{tabular}
\caption{ A detailed comparison between \mval and \mgraph on the results of our main experiments for $\real$ and $\synth$, where we average MSE ($\pm$ standard deviation) over all leave-1-out test samples. We find that \mgraph has a generally consistent improvement over \mval.}
\label{tbl:loo_val_vs_graph}
\end{table}

\begin{table}[htbp]
\centering
\begin{tabular}{ l r c c }
Scalar Name & \tworow{Kustowski}{et al.$(\real_7)$} & $\boldsymbol{\theta}^* (\real_7)$ & $\boldsymbol{\theta}^s (\real_7)$ \\ \hline
Neutron bang time      & 8.822{\scriptsize$\pm 2.779$ }& 0.049{\scriptsize$\pm 0.053$ }& 0.042{\scriptsize$\pm 0.037$ }\\ 
X-ray bang time        & 10.219{\scriptsize$\pm 3.305$ }& 0.053{\scriptsize$\pm 0.078$ }& 0.047{\scriptsize$\pm 0.042$ }\\ 
Downscattered ratio    & 121.336{\scriptsize$\pm 13.337$ }& 0.694{\scriptsize$\pm 0.358$ }& 0.670{\scriptsize$\pm 0.382$ }\\ 
Temperature            & 87.385{\scriptsize$\pm 12.098$ }& 0.170{\scriptsize$\pm 0.113$ }& 0.178{\scriptsize$\pm 0.102$ }\\ 
Hot spot radius        & 11.430{\scriptsize$\pm 3.723$ }& 0.122{\scriptsize$\pm 0.083$ }& 0.157{\scriptsize$\pm 0.104$ }\\ 
Velocity               & 137.237{\scriptsize$\pm 28.350$ }& 0.391{\scriptsize$\pm 0.325$ }& 0.271{\scriptsize$\pm 0.199$ }\\ 
X-ray emission         & 229.221{\scriptsize$\pm 26.436$ }& 4.913{\scriptsize$\pm 22.881$ }& 1.264{\scriptsize$\pm 1.175$ }\\ 
Neutron yield          & 81.705{\scriptsize$\pm 19.727$ }& 0.134{\scriptsize$\pm 0.177$ }& 0.065{\scriptsize$\pm 0.103$ }\\ 
Neutron burn width     & 2.880{\scriptsize$\pm 0.579$ }& 0.340{\scriptsize$\pm 0.241$ }& 0.324{\scriptsize$\pm 0.218$ }\\ 
X-ray burn width       & 44.148{\scriptsize$\pm 9.635$ }& 3.405{\scriptsize$\pm 2.098$ }& 3.293{\scriptsize$\pm 2.185$ }\\ 
Average (above scalars)& 73.438{\scriptsize$\pm 6.167$ }& 1.027{\scriptsize$\pm 1.612$ }& 0.631{\scriptsize$\pm 0.957$ }\\ \hline
Images                 & 1.445{\scriptsize$\pm 0.221$ }& 0.208{\scriptsize$\pm 0.081$ }& 0.189{\scriptsize$\pm 0.068$ }\\ \hline
\end{tabular}
\caption{The average MSE over all leave-three-out test samples using our graph optimized model, compared to the baseline, on both the simulated and experiment datasets. Our model often has large performance increases over the baseline for both scalar predictions and image predictions. }
        \label{tbl:leave3out}
\end{table}

\begin{table}[htbp]
\centering
\begin{tabular}{lcccc}
Scalar Name                     & Baseline Mean (Std) & Val Mean (Std) & Graph Mean (Std) \\ \hline
Neutron bang time                & 0.2432 (0.3485) & 0.0482 (0.0678) & \textbf{0.0367 (0.0408)} \\
X-ray bang time                  & 0.2667 (0.3159) & 0.0363 (0.0532) & \textbf{0.0293 (0.0269)} \\
Downscattered ratio              & 0.9203 (1.2813) & 0.7054 (0.7414) & \textbf{0.5501 (0.5407)} \\
Temperature                      & 0.2331 (0.2437) & 0.1748 (0.1782) & \textbf{0.1516 (0.1362)} \\
Hot spot radius                  & 0.1302 (0.1928) & \textbf{0.1096 (0.1393)} & 0.1155 (0.1245) \\
Velocity                         & 0.3214 (0.2945) & 0.3310 (0.3873) & \textbf{0.2122 (0.2180)} \\
X-ray emission                   & 1.3628 (1.8793) & 1.2416 (1.8177) & \textbf{0.7455 (1.0990)} \\
Neutron yield                    & 0.0577 (0.0401) & 0.1453 (0.1381) & \textbf{0.0347 (0.0381)} \\
Neutron burn width               & 0.4040 (0.5783) & \textbf{0.2694 (0.2782)} & 0.3202 (0.4238) \\
X-ray burn width                 & 4.7584 (7.0234) & 2.7827 (4.5374) & \textbf{2.7282 (3.9024)} \\
Average                          & 0.9118 (1.0399) & 0.5844 (0.8134) & \textbf{0.4924 (0.7787)} \\ \hline
Images                             & 0.1700 (0.1278) & 0.1583 (0.1021) & \textbf{0.1535 (0.0948)} \\ \hline \hline
\end{tabular}
\caption{The average MSE over all leave-1-out test samples using our surrogate and graph on $\real$.}
\label{tbl:results_leave1_out_real}
\end{table}

\begin{table}[tb]
\centering
\begin{tabular}{lcccc}
Scalar Name                    & baseline Mean (Std) & val Mean (Std) & graph Mean (Std) \\ \hline
Neutron bang time             & 0.8039 (0.1922)     & \textbf{0.6355 (0.2302)} & 0.6641 (0.2756) \\
X-ray bang time               & 1.0368 (0.2753)     & \textbf{0.6466 (0.2008)} & 0.6793 (0.2590) \\
Downscattered ratio           & 5.4904 (1.4794)     & 5.8749 (3.5769) & \textbf{4.4946 (1.7533)} \\
Temperature                   & 4.3514 (0.7943)     & \textbf{2.8154 (0.9657)} & 2.8925 (0.7221) \\
Hot spot radius               & 9.0593 (1.6350)     & 7.1869 (1.6696) & \textbf{6.7878 (1.6811)} \\
Velocity                      & 8.6147 (1.9951)     & 7.6575 (3.8849) &\textbf{ 6.9695 (2.7580)} \\
X-ray emission                & 8.2617 (1.8245)     & \textbf{4.1908 (1.9434)} & 4.5162 (1.9517) \\
Neutron yield                 & 8.3890 (1.8586)     & \textbf{3.8138 (1.6989)} & 4.3552 (1.7640) \\
Neutron burn width            & 9.0299 (1.6617)     & 10.0658 (3.1624) & \textbf{8.8512 (4.5460)} \\
X-ray burn width              & 10.7740 (1.9504)    & 12.7686 (4.5481) & \textbf{11.3418 (5.4129)} \\
Average                       & 6.5811 (0.5218)     & 5.5656 (3.7451) & \textbf{5.1552 (3.2357)} \\ \hline
Images                        & 0.0789 (0.0070)     & 0.0315 (0.0088) &\textbf{ 0.0304 (0.0095)} \\ \hline \hline
\end{tabular}
\caption{The average MSE over all leave-1-out test samples using our surrogate and graph on $\synth$.}
\label{tbl:results_leave1_out_sim}
\end{table}
